\pdfoutput=1

\documentclass[11pt]{article}

\usepackage[]{emnlp2021}

\usepackage{times}
\usepackage{latexsym}
\usepackage{url}
\usepackage{latexsym}
\usepackage{amsmath}
\usepackage{booktabs}
\usepackage{graphicx}

\usepackage{mathtools}
\usepackage{amsmath, bm}
\usepackage{listings} 
\usepackage{color}
\usepackage{subfig}
\usepackage{multirow}
\usepackage[normalem]{ulem}
\useunder{\uline}{\ul}{}
\usepackage{float}

\usepackage[T1]{fontenc}

\usepackage[utf8]{inputenc}

\usepackage{microtype}

\title{Informed Sampling for Diversity in Concept-to-Text NLG}

\author{Giulio Zhou 
  \\Huawei Noah's Ark Lab\\ London, UK\\ \texttt{giuliozhou@huawei.com} \\\And
  Gerasimos Lampouras 
  \\Huawei Noah's Ark Lab\\ London, UK\\ \texttt{gerasimos.lampouras@huawei.com } \\}
\begin{document}\maketitle
\begin{abstract}
Deep-learning models for language generation tasks tend to produce repetitive output. Various methods have been proposed to encourage lexical diversity during decoding, but this often comes at a cost to the perceived fluency and adequacy of the output. In this work, we propose to ameliorate this cost by using an Imitation Learning approach to explore the level of diversity that a language generation model can reliably produce. Specifically, we augment the decoding process with a meta-classifier trained to distinguish which words at any given timestep will lead to high-quality output. We focus our experiments on concept-to-text generation where models are sensitive to the inclusion of irrelevant words due to the strict relation between input and output. Our analysis shows that previous methods for diversity underperform in this setting, while human evaluation suggests that our proposed method achieves a high level of diversity with minimal effect to the output's fluency and adequacy.

\end{abstract}

\section{Introduction}

\begin{figure*}[tbp]

\centering

  \includegraphics[width=\linewidth]{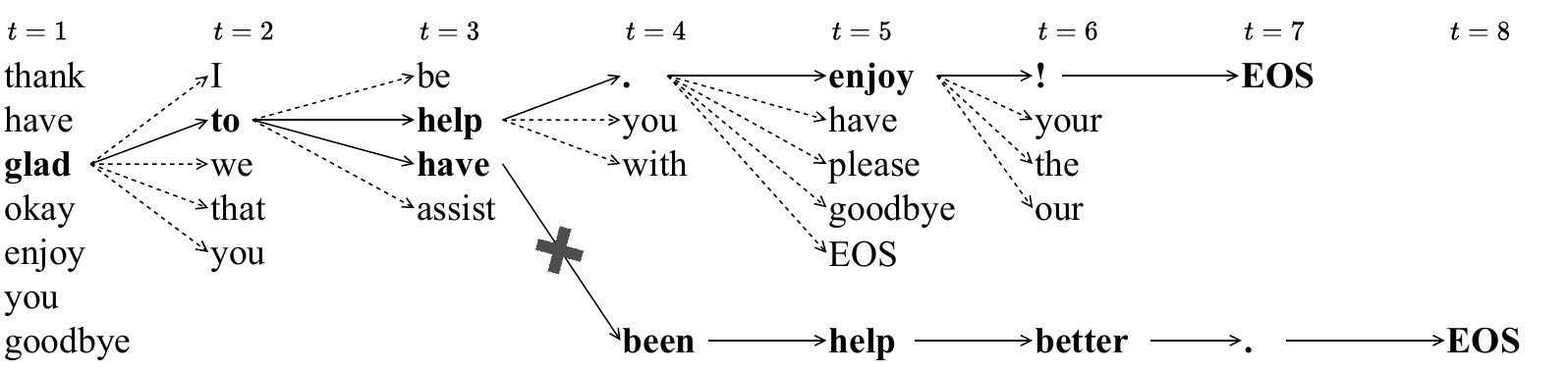}

\caption{Example decoding for \textsc{[inform(welcome); inform(bye)]}; diverging at the third time step.}
  \label{fig:varietyexample}

\end{figure*}

The use of deep-learning models for language generation tasks has become prevalent in recent years 
as they achieve high performance without manually engineered rules or features
\cite{wen2015sclstm, mei2016talk, duvsek2018findings}. 
However, while the produced texts are qualitatively acceptable according to most evaluation criteria, they are often repetitive or disfluent when multiple diverse outputs are needed. This problem is attributed to using the maximum-likelihood objective function for training as it encourages the generation of highly frequent words and sentence structures, i.e. models overfit and do not learn to exploit the lexical and structural diversity that is present in the dataset \citep{li2016diversity}.

Here we focus on concept-to-text Natural Language Generation (\textsc{nlg}), where the input is a meaning representation (\textsc{mr}) and the output is an utterance expressing the input in natural language. Due to the stricter relation between input and output, it is more challenging to promote diversity in concept-to-text than other language generation tasks. Diverging from greedy inference can lead to error propagation that negatively affects the output's relevance to the input. However, assuming the output is sequentially decoded, most research on concept-to-text diversity focuses on sampling over the probability distribution \citep{wen2015sclstm}.

More complex decoding strategies have been proposed for the related task of open-domain \textsc{nlg}, where the input is a natural language context and the output is a relevant response. \citet{fan2018hierarchical} limit the decoding distribution to a fixed number of the Top-$k$ words (Top-$k$ Sampling), while \citet{holtzman2019curious} limit the distribution to the largest subset of words whose 
cumulative probability does not exceed a predefined parameter $p$ (Nucleus Sampling). Nucleus Sampling improves over Top-$k$ by retaining a dynamic number of words per decoding step, but the probability mass $p$ remains a constant parameter.
However, these strategies are sensitive to their parameters $k$ and $p$ and there is no established methodology to tune them so that the output fluency and adequacy do not suffer while also achieving high diversity. 

In this paper, we propose \textit{Informed Sampling} for diversity, i.e. to sample amongst \textit{reliable} words that lead to diverse output but are not liable to lead to disfluent word sequences through error propagation. To distinguish which words in the decoding distribution can be reliably sampled by the \textsc{nlg} model, we employ a meta-classifier 
that leverages a diversity-specific training signal. 
Our approach is only applied during decoding and is orthogonal to the architecture of the \textsc{nlg} model, which we assume as pretrained. Unlike previous decoding strategies, \textit{Informed Sampling} 
does not depend on manually tuned parameters. 


As there is no explicitly annotated data for \textit{Informed Sampling}, we adapt three Imitation Learning (IL) frameworks to train the meta-classifier; IL is a family of meta-learning frameworks that train models based on expert demonstrations. We design an expert policy as to infer which words are reliable based on what the \textsc{nlg} model can produce without negatively affecting the output's quality. 
Through this, the meta-classifier is fitted to the level of diversity captured by the \textsc{nlg} model.

We present experimental analysis on the application of IL to the meta-classifier and compare against related approaches.
Additionally, this paper explores the concept-to-text application of diversity methods originally proposed for open-domain \textsc{nlg}. 
Automatic and human evaluation suggests that \textit{Informed Sampling} produces diverse output while maintaining its fluency and adequacy.

\section{Related Work}

There have been a number of different approaches to encourage output diversity in open-domain \textsc{nlg}. \citet{li2016diversity} propose mutual information maximization as a diversity focused objective function, while \citet{zhang2018generating} propose variational information maximization in combination with adversarial learning.
\citet{zhao2017learning} produce diverse output by augmenting the input encoding with diversity-specific information through Conditional Variational Autoencoders. Going further with modifying the encoding, \citet{gao2019jointly} reshape the whole embedding space of the input, arguing that a more structured latent space leads to more diverse output. 
We explore the application of these methods to concept-to-text \textsc{nlg} in later sections, but we find that they underperform compared to their open-domain use. These methods promote semantic diversity, and might be incompatible with concept-to-text where the output semantics are strictly bounded by the input. 

Research on neural output diversity for concept-to-text \textsc{nlg} is limited and mostly focused on different decoding strategies (e.g. beam search). 
Most recently, \citet{deriu2017end} proposed ``forcing'' the output of the first decoding step, arguing that greedy inference from different starting points leads to diverse but fluent sentences. They achieve this by augmenting the input to bias the first step of the decoding process towards particular words observed in the data. However, the application of their method is limited to the first decoding step.


Imitation Learning frameworks have been applied on a variety of structured prediction NLP tasks, such as dependency \cite{goldberg-nivre-2013-training} and semantic parsing \cite{vlachos-clark-2014-new}. Most related to this work, the \textsc{lols} framework was applied to concept-to-text \textsc{nlg} from unaligned data \cite{lampouras-vlachos-2016-imitation}.


\section{Meta-Classifier for Diversity}


Concept-to-text \textsc{nlg} is the task of converting a machine-interpretable \textsc{mr} into natural language text. The input \textsc{mr} consists of one or more predicates; each predicate has a set of attributes and corresponding values. The predicate dictates the communication goal of the output text, while attributes and values dictate content. 
For example, the \textsc{mr} \textsc{[inform(rest-name = Mizushi, Okasan)]} denotes that the output should inform the user of two restaurants called ``\textit{Mizushi}'' and ``\textit{Okasan}''.
Concept-to-text datasets
usually provide multiple output references per \textsc{mr}. Specifically, the MultiWOZ dataset \citep{budzianowski2018multiwoz} provides 1872 distinct references for the \textsc{mr} \textsc{[inform(welcome); inform(bye)]}, e.g. ``\textit{Glad to help. Enjoy!}'', ``\textit{Glad to assist you. Goodbye.}'' 

We treat \textsc{nlg} as a structured prediction problem, where the output is a sequence of words constructed via sequential decoding. 
Informed Sampling is orthogonal to the architecture of the \textsc{nlg} model, only assuming a sequential decoding process.
Figure \ref{fig:varietyexample} shows a partial example of the diversity exhibited by a trained \textsc{nlg} model, for the previously mentioned \textsc{mr}. At each timestep we can examine the distribution that results from decoding and sample accordingly to promote diversity; the words are shown in descending probability. However, only a subset of words in the vocabulary will lead to fluent and adequate sequences. As mentioned before, we denote these as \textit{reliable words}. For example, in $t=3$ choosing the word ``\textit{have}'' seems like a sensible choice given the history; one can imagine that this may lead to an output like ``\textit{glad to have been of help!}'' Unfortunately, due to the \textsc{nlg} model being imperfect, this will actually lead to the disfluent output ``\textit{glad to have been help better.}'' On the contrary, the word ``\textit{assist}'' has less probability than ``have'' but it leads to the same 
subtree as ``\textit{help}'', and to fluent output.

\begin{figure}[htb]
\centering
  \includegraphics[width=\linewidth]{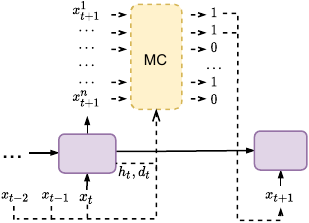}
\caption{Overview of the meta-classifier (MC); dotted lines denote the MC, solid denotes \textsc{nlg} model.}
  \label{fig:meta}
\end{figure}

As briefly mentioned in the introduction, we propose to use a meta-classifier (see Figure~\ref{fig:meta}), external to the \textsc{nlg} model, that learns to distinguish which words in the decoding distribution are \textit{reliable}. 
The meta-classifier is a simple feed-forward neural network composed of alternating linear and ReLU layers ending with a softmax. It considers each word in the \textsc{nlg} model vocabulary individually, and the output for each is a probability distribution over values 0 and 1; 1 denotes the word as \textit{reliable}.

The input $\textbf{c}$ for a given word is a concatenation of 
the \textsc{nlg} states and embeddings (eq. \ref{eq:input}).
\begin{equation}\label{eq:input}
\begin{split}
\textbf{c} = & [\textbf{h}_{t}, W_{dc}\textbf{d}_{t}, W_{wr}\textbf{x}^{i}_{t+1}, \\ & W_{wr}\textbf{x}_{t-2}, W_{wr}\textbf{x}_{t-1}, W_{wr}\textbf{x}_t] \\
\end{split}
\end{equation} 
\noindent
where ${h}_{t}$ is the hidden state at step $t$, $W_{dc}$ is the input representation weight matrix, ${d}_{t}$ represents input to be generated, $W_{wr}$ is a word embedding weight matrix, ${x}_{t}$ is the word at step $t$, and ${x}^{i}_{t+1}$ is the $i$-th word of the decoding distribution at $t+1$. 
In this paper we use notations specific to the SCLSTM  architecture \citep{wen2015sclstm}. However, the input of the meta-classifier can be generalised as a concatenation of encoder, decoder hidden states and word embeddings.

From the meta-classifier's output we can infer a vocabulary-length binary vector $B$ that indicates which words are \textit{reliable}. In order to also consider the \textsc{nlg} decoder's probability distribution, we only sample amongst the top consecutive \textit{reliable words} in $B$ that are assigned a non-zero probability. 

\section{Imitation Learning} \label{IL}

Since diversity-specific labels are not explicitly available in the data, we employ an expert policy to infer which words are \textit{reliable}, and use Imitation Learning (IL) approaches to mimic the expert. IL is a family of meta-learning frameworks, that train a policy $\pi$ using demonstrations provided by an expert $\pi^{ref}$. In this work, the policy $\pi$ refers to the meta-classifier. The expert policy $\pi^{ref}$ acts as a dynamic oracle that returns whether a word is \textit{reliable}; we discuss the expert further in section~\ref{expert}. 

We explore the application of three IL frameworks for training the meta-classifier: Exact Imitation, \textsc{DAgger} \citep{ross2011reduction} and Locally Optimal Learning to Search \citep[LOLS]{chang2015learning}. We will briefly explain how we adapt these frameworks, but a detailed explanation of the involved algorithms is out of the paper's scope.



Exact Imitation refers to training a policy $\pi$ directly on the labels provided by the expert policy $\pi^{ref}$. In practice, for each training instance in our data, we use the underlying \textsc{nlg} model to generate a sentence. On each decoding step, we call $\pi^{ref}$ to determine the \textit{reliable} words and train $\pi$. We also sample the next word in the sentence using $\pi^{ref}$. We note that the underlying \textsc{nlg} model remains constant throughout training and IL is applied solely on the meta-classifier.

\textsc{DAgger} improves over Exact Imitation by generating the sentence using a mixture of the $\pi^{ref}$ and $\pi$ policies, i.e. by sampling amongst the words considered \textit{reliable} by either $\pi^{ref}$ or $\pi$. This way $\pi$ is exposed to sentences it would not have encountered solely using $\pi^{ref}$ for sampling. In particular, it is exposed to sentences produced by $\pi$ itself, in a sense exposing it to its own errors and thus helping ameliorate error propagation. As in Exact Imitation, we call $\pi^{ref}$ to determine \textit{reliable} words and train $\pi$. 
Before we apply \textsc{DAgger}, we perform one iteration of Exact Imitation to initialise $\pi$.

LOLS generates the sentence using only the $\pi$ policy, again initialised via Exact Imitation. Additionally, at each decoding step the training signal is provided by either $\pi^{ref}$ or $\pi$ according to a probability $p$. This probability is initially set to $p=1.0$, i.e. to always obtain the training signal via $\pi^{ref}$, but it exponentially decays after every iteration with a rate of $p = (1-\beta)^i$, where $\beta$ is the learning rate. Further details on how $\pi$ can provide a training signal can be found in section~\ref{expert}. 

\textsc{DAgger} and \textsc{lols} iteratively adjust the training signal and increasingly expose $\pi$ to training instances that are more similar to what $\pi$ is likely to encounter during test time. This helps address error propagation, but also helps tune the meta-classifier to the level of diversity that the \textsc{nlg} model can comfortably produce. Specifically, \textsc{lols} has the advantage of potentially improving over $\pi^{ref}$ as it exploits the training signal from $\pi$ itself.

\subsection{Inferring Training Signal from Policies}\label{expert}

\begin{table}[tb]
\resizebox{\columnwidth}{!}{\begin{tabular}{lll|l|c}
\multicolumn{1}{l|}{\begin{tabular}[c]{@{}l@{}}${x}_{t}$\end{tabular}} & \multicolumn{1}{l|}{${x}^{i}_{t+1}$} & Greedy decoding              & $Prec^{i}$ & Out \\ \hline
my                                                                       & favourite                       & is Itacho . do       & 0.908         & 1     \\
my                                                                       & personal                       & favorite is Itacho . & 1.0         & 1     \\
my                                                                       & recommendation                 & is the hotel Hilton   & 0.524     & 0    \\
my                                                                       & computer                       & shows this . I        & 0.658       & 0     \\
my                                                                       & opinion                        & . I would recommend  & 0.708       & 0     \\
my                                                                       & suggestion                     & would be the Itacho  & 1.0         & 1     \\
my                                                                       & apologies                      & . I would suggest    & 0.608       & 0    
\end{tabular}}
\caption{Example of $\pi^{ref}$ training signal inference.}
\label{tab:expert}
\end{table}






During IL, we employ a dynamic oracle $\pi^{ref}$ that determines whether a word ${x}^{i}_{t+1}$ is \textit{reliable}. Due to the computational cost, $\pi^{ref}$ is limited to consider only $i \in \{0...d\}$; in this work we consider the top $d=25$ words, which is the maximum number of consecutive \textit{reliable} words as observed during preliminary training. This limit is not applied during decoding with the trained meta-classifier. 

Intuitively, we need to examine whether the impact of each ${x}^{i}_{t+1}$ on the decoding process will lead to a fluent and adequate sentence. To obtain sentences that are affected by ${x}^{i}_{t+1}$, we force ${x}^{i}_{t+1}$ in step $t+1$ and use the \textsc{nlg} model to greedily generate the rest of the sentence. We then calculate the n-gram overlap between the $d$ sentences and a set of references. To make the calculations more consistent, we limit the produced sentences to the previous word ${x}_{t}$, ${x}^{i}_{t+1}$, and the next 4 words ${x}^{*}_{t+2} \dots {x}^{*}_{t+5}$, similarly to the focused costing approach proposed by \citet{goodman-etal-2016-noise}. If a sequence ends prematurely (e.g. by generating an $\langle eos \rangle$ token), we pad it to the appropriate length.

An example application of $\pi^{ref}$ is shown in Table~\ref{tab:expert} for the \textsc{mr} \textsc{[inform(rest-name = itacho), request(rest-type)]}. Note that the previous word ${x}_{t}$ is the same for all examined ${x}^{i}_{t+1}$, while ${x}^{*}_{t+2} \dots {x}^{*}_{t+5}$ differ. The n-gram overlap is calculated via modified 4-gram precision, i.e. BLEU-4 score \cite{papineni2002bleu} without the brevity penalty. Since the expert hypotheses are all fixed in size, we cut the brevity penalty to speed up the calculation of the expert. 
The expert considers the words and corresponding modified 4-gram precisions $Prec^{i}$ in ascending $i$, considering a word $i$ as \textit{reliable} if $Prec^{i} \geq max(Prec^{0}, \dots, Prec^{i-1})$. 



To promote more diversity through $\pi^{ref}$, the aforementioned reference sets are obtained by decomposing the corresponding \textsc{mr} into its attributes, and then retrieving from the training instances all the references these attributes correspond to. For example, for \textsc{[inform(welcome); inform(bye)]} we would also retrieve all references corresponding to \textsc{[inform(welcome); request(name)]} as they share the \textsc{inform(welcome)} attribute.

In the \textsc{lols} framework, we also obtain the training signal via $\pi$. In this work, this is similar to how we calculate $\pi^{ref}$ but instead of greedily generating the rest of the sentence for each ${x}^{i}_{t+1}$, we generate by sampling using $\pi$.
In order to allow a broader exploration and generate a more consistent signal when sampling, multiple hypotheses are produced and precision is averaged over them.

\section{Experiments}


The following experimental analysis is performed on the MultiWOZ dataset \cite{budzianowski2018multiwoz} which contains human-to-human written conversations, annotated with corresponding \textsc{mr}s.
The conversations concern a user trying to use a virtual assistant to perform certain tasks, e.g. book a restaurant or a taxi, find attractions. The dataset is comprised of 55026, 7290 and 7291 utterances for training, validation and testing respectively. In the training set, there are 486 different attributes, 8635 unique \textsc{mr}s and a total of 46671 distinct sentences. We note that scarcity of data is one of the major challenges of concept-to-text \textsc{NLG}, and that MultiWOZ is one of the largest and more diverse datasets available. Both DSTC8 and DSTC9 challenges use MultiWOZ in their tasks\footnote{https://sites.google.com/dstc.community/dstc8/tracks, https://sites.google.com/dstc.community/dstc9/tracks}.

 \subsection{Evaluation Metrics}\label{evalMetric} 
To measure the diversity of the outputs we compute Self-BLEU \citep{zhu2018texygen} and diversity-n \cite{li2016diversity}. 
In our experiments, we will be reporting 1 - Self-BLEU to make the score easily interpretable (the higher the score, the more diverse the system output is), while for distinct-n we provide 
the percentage of distinct n-grams ($n={1,2,4}$)
and distinct whole sentences.

Correctness of the output is evaluated with slot error \cite[ERR]{wen2015stochastic}, i.e. the percentage of values in the \textsc{mr} that are missing, repeated or hallucinated in the output. Overall performance is evaluated with BLEU-4, METEOR \cite{lavie2007meteor} and MoverScore \cite{zhao2019moverscore}. We should note that word overlap metrics can be unreliable when evaluating systems with a high level of diversity in the output. Since every \textsc{mr} is aligned with a limited set of references, more diversity will lead to less overlap between the output and the references. BLEU is particularly problematic, as it has been shown not to be a reliable discriminator between high quality systems even when not considering a particularly diverse output \cite{novikova-etal-2017-need}. For this reason, we further support our experiments with human evaluation.

\subsection{System Configurations} \label{configuration}

Apart from the experiments with SpaceFusion \cite{gao2019jointly}, all our experiments make use of the Semantically Conditioned Long Short-term Memory (SCLSTM) architecture, proposed by \citep{wen2015sclstm}, as the underlying \textsc{nlg} model. While recent architectures have been adapted to take advantage of large pretrained language models \citep{peng2020few-shot}, we opt not to use them here as related work does not exploit external data either.
For our meta-classifier, we initialised using a single iteration of Exact Imitation over the full dataset. 
Due to time constraints, for the following training iterations with any IL framework, only 10\% randomly selected sentences were used. We evaluated the meta-classifiers generated by our last iteration. 
For the MMI objective function we implemented MMI-antiLM as suggested by \citet{li2016diversity}. At decoding, the MMI-antiLM parameters were set as $\lambda=0.5$ and $g=5$. 
Beam Search and MMI are performed with beam size = 10.
SpaceFusion 
was trained using the configuration provided with the code. Tests on different settings did not achieve significant improvements. Values for the random vector $r$ were generated in the range $-5, 5$.
For First Word Control, we selected all the words that appear more than 60 times as first word in the training references, resulting in a set of 67 different possible first words. At inference time, one sentence is generated per each first word.
For Top-$k$ and Nucleus Sampling, since parameters $k$ and \textit{p} are not tunable, we report results for ranges 2-10 for $k$ and 0.10-0.95 for $p$. 

The aforementioned parameters in related work ($\lambda$, $g$, $r$, $k$, $p$), were all tuned based on observations of output and diversity metrics. Precise tuning of such manual parameters remains a challenge as word-overlap metrics are unreliable predictors of actual output quality (see Section~\ref{evalMetric}). 

\subsection{Reranking}\label{output} 
For each input, we generate 10 possible outputs and rerank them according to two criteria. We prioritize utterances with lowest slot error, and then sort them according to their normalised sentence probability. The final output is sampled uniformly from the top 5 most probable remaining sentences. This is applied on all considered models to minimise the effect of random sampling on the results.


\begin{table*}[tb]
\centering
\resizebox{\textwidth}{!}{
\begin{tabular}{l|cccccccccc}
           & Greedy               & Beam                 & FWC                   & MMI                   & SF     & Top-$k$ & Nucleus     & IS-E & IS-D        & IS-L     \\ 
           &                &   \small{$b = 10$}      &             &        \small{$g,\lambda=5,0.5$}     &   \small{$|r| = 5$}   & \small{ $k = 2$} &\small{ $p = 0.84$}     &  &         &      \\ \hline
BLEU       & 0.654                & \textbf{0.663}                & 0.592               & 0.486                & 0.439  & 0.336 & 0.488     & 0.326  & 0.334  & 0.334              \\
METEOR      & \textbf{0.496}  & \textbf{0.496} & 0.479  &  0.479 & 0.332 & 0.400 & 0.434 & 0.393  & 0.395  & 0.395\\
Mover      & \textbf{0.804}                & 0.799                & 0.721                 &      0.649                &          0.642  & 0.675 & 0.710  & 0.646  & 0.645 & 0.649        \\
Slot Error & 4.071                & 1.608                & \textbf{0.305}                 &        2.091              &     45.218  & 0.830 & 0.753    & 0.897  & 0.762  & 0.897            \\ \hline
1-SB       & 0.014                & 0.017                & 0.018                 &                  0.044    &  0.008  & 0.093 &0.101     & \textbf{0.104}  & 0.096  & 0.096             \\
Dist-1     & 0.004  & 0.004 & 0.004  & 0.006 & 0.002 & 0.007 & \textbf{0.007} & \textbf{0.007}  & \textbf{0.007}  &\textbf{0.007} \\
Dist-2     & 0.022  & 0.024 & 0.023 & 0.049 & 0.013 & 0.066 & \textbf{0.072} &  0.064  & 0.061  & 0.061\\
Dist-4      & 0.079  & 0.087 & 0.095 & 0.156 & 0.045 & 0.399 & 0.342 & \textbf{0.429}  & 0.415  & 0.417  \\
Dist-Sent  & 0.307                & 0.482                & 0.491                 &          0.487            &  0.266  &0.869 &0.919 & \textbf{0.961}  & 0.957  & 0.957 
\end{tabular}}
\caption{Automatic evaluation results for different methods on diversity. 
IS-X refers to Informed Sampling trained with either Exact Imitation (IS-E), \textsc{DAgger} (IS-D) or \textsc{lols} (IS-L).
}
\label{tab:results}
\end{table*}

\subsection{Analysis of Previous Diversity Methods}\label{sec:previous}

Please consult Table~\ref{tab:results} for automatic evaluation metrics. 
We can see from the low numbers in the diversity metrics (1 - Self-BLEU and distinct-n) that none of previous diversity methods produce much output diversity in concept-to-text NLG.
Below, we provide some brief analysis on the results.

\noindent
\textbf{Beam Search:} similarly to what has been reported in open-domain \textsc{nlg} research \cite{li2016diversity}, Table \ref{tab:results} shows that Beam Search produces greedy-like outputs with minimal variations. 

\noindent
\textbf{MMI-AntiLM}: using Beam Search with MMI as objective function improved the diversity of the output. However, an analysis of the text revealed that the generated sentences do not differ substantially from the ones obtained with maximum-likelihood, and that the achieved diversity was the result of introducing disfluent words within the first $g$ tokens.




\noindent
\textbf{First Word Control (FWC):} despite being proposed for concept-to-text \textsc{nlg}, First Word Control did not achieve more diversity than Beam Search. 
We have observed that in most cases the forced first word has no major effect on the sentence. For example, for the \textsc{mr} \textsc{[request(taxi-leave)]}, forcing ``\textit{Okay}'', ``\textit{Alright}'' or ``\textit{Great}'' will not produce diversity as the model will complete the sentence with ``\textit{What time would you like to leave?}''. 
  


\begin{figure}%
    \centering
   \includegraphics[width=7.7cm]{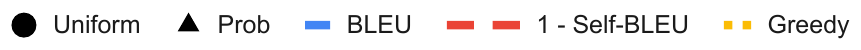} %
   
    \subfloat[Top-$k$]{{\includegraphics[width=3.7cm]{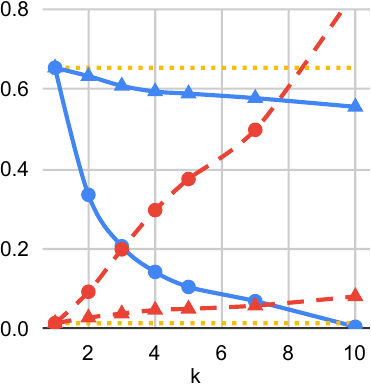} }\label{fig:k}}%
     \subfloat[Nucleus Sampling]{{\includegraphics[width=3.8cm]{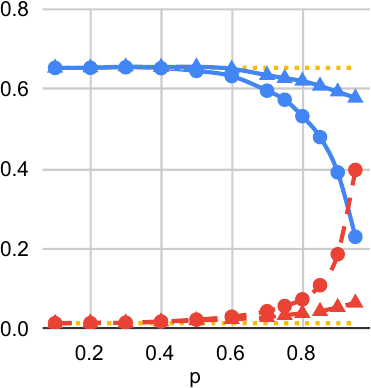} }\label{fig:nuc}}%
    \caption{BLEU and 1 - Self-BLEU for Nucleus Sampling / Top-$k$ with uniform / stochastic sampling.}%
    \label{fig:nucleus}%
\end{figure}

\begin{figure}%
    \centering
    \subfloat[Average prob. distribution]{{\includegraphics[width=3.7cm]{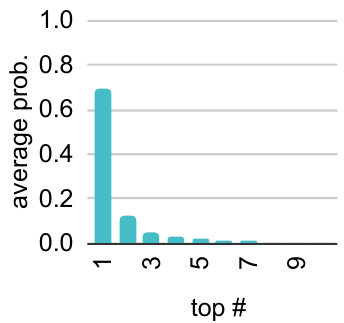} }\label{fig:top10}}%
     \subfloat[Probability of the top-1]{{\includegraphics[width=3.8cm]{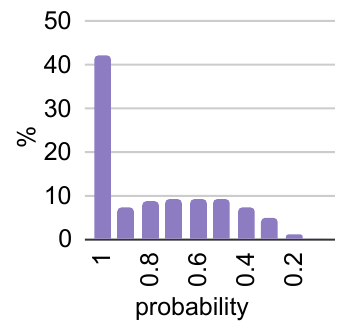} }\label{fig:first}}%
    \caption{Close up of the distribution produced by the concept-to-text \textsc{nlg} model.}%
    \label{fig:dist}%
\end{figure}
\noindent
\textbf{SpaceFusion (SF):} compared to the above methods SpaceFusion obtained the lowest scores for diversity and highest slot error. This makes sense as the method was not designed for concept-to-text \textsc{nlg} nor for lexical diversity in general, and the trained autoencoder tends to produce identical or almost identical sentences as the S2S encoder. The joint training collapses the sentence embeddings into a similar representation, preventing the decoder from distinguishing different autoencoder states. This is explained by the strict semantic relation between the \textsc{mr} and the reference, and the similarities within the reference set and the fuse regularisation term. The extremely high slot error can be attributed to the lack of an attention mechanism in SpaceFusion. Widening the range of the random vector added to the latent vectors can increase diversity but not without reducing the relevancy of the sentences further. We conjecture that SpaceFusion might achieve a better performance with an input-optimised model and parameters, but that is beyond the scope of this paper.

Figure \ref{fig:nucleus} show how the quality of the texts produced by Top-$k$ and Nucleus Sampling when paired with stochastic sampling vary as their respective parameters increase.\footnote{We present detailed results in the Appendix.} Despite enlarging the sample pool results in the augmentation of diversity, Top-$k$ and Nucleus Sampling performed comparably across all the parameters, obtaining greedy-like results. Figure \ref{fig:top10} shows the average probability of the top-10 words. Since most of the probability mass is clustered in the top 4 words, with the top-1 taking 70\% of it, we can conclude that stochastic sampling is not appropriate for concept-to-text \textsc{nlg} as little to no diversity would be introduced.

\subsection{Top-$k$ and Nucleus Sampling Analysis}\label{nucvsk}

In addition to stochastic sampling, Figure \ref{fig:nucleus} shows the performance of Top-$k$ and Nucleus Sampling when paired with uniform sampling. For Top-$k$ (Figure \ref{fig:k}), while the diversity in the text increases drastically, the BLEU score drops exponentially over $k$. The score halves even for $k=2$ (one step beyond greedy decoding) and reaches a 0.005 BLEU score at $k = 10$. Figure \ref{fig:first} shows that 42\% of the generated words have a probability of 1.0 (or nearly 1.0). Even though diversity methods aim to reduce the bias towards highly probable words, it is safe to assume that  in concept-to-text \textsc{nlg}, words with a probability of 1.0 are likely to be the sole correct output. For this reason, when $k$ increases, errors on these cases become more probable. In addition, it is fairly reasonable to trust the low scores of word-overlap metrics on the incorrectness of the output produced by \textit{top-k} due to their high correlation with human judgements when evaluating low-quality text \cite{zhao2019moverscore}.

On the other hand, Nucleus Sampling paired with uniform sampling is able to introduce diversity in a more controlled way, outperforming Top-$k$ by maintaining a high level of BLEU while steadily increasing the diversity generated. (Figure \ref{fig:nuc}). 
We note that Nucleus Sampling can achieve any desired level of diversity through different values of $p$.
However, picking an optimal value for $p$ is not straightforward as the effect of each level of diversity to the quality of the output is unreliably measured by the word overlap metrics.



\subsection{Evaluation of Informed Sampling} \label{sec:evalis}

Table \ref{tab:results} also shows our three Informed Sampling models trained with Exact Imitation (IS-E), \textsc{DAgger} (IS-D) and \textsc{lols} (IS-L). 
All the configurations obtained comparable automatic evaluation results, suggesting that the benefits of \textsc{lols} do not help in this task. We conjecture this is due to the high quality of the expert policy which provides a reliable and representative training signal for the diversity that the \textsc{nlg} is capable of producing correctly. 

Compared to previous methods our approaches show a much higher level of diversity in the output. However, we observe a significant drop in the word overlap metrics (BLEU-4, METEOR and MoverScore). As we mentioned in section~\ref{evalMetric}, these metrics rely on a limited set of evaluation references, and are unfortunately unreliable when there is a high level of diversity in the output. We consider Self-BLEU and distinct-n to be accurate as they do not rely on references. To better determine the output's quality, we perform human evaluation.

For human evaluation we include the output of Nucleus Sampling and greedy decoding. To further focus the human evaluation solely on output quality, we aim to keep the level of diversity across systems as close as possible.
The behavior of greedy decoding is not adjustable, but we can adjust the level of diversity of Nucleus Sampling using different $p$ values.
Unfortunately, we cannot use the development data to pick $p$ as we observed it leads to different Self-BLUE values in the test set which would compromise the comparison.
We set $p = 0.84$ as that leads to the same Self-BLEU as our systems on the test data. 
This also leads to a higher BLEU score by 0.16 points, but the difference for semantic similarity based metrics is more marginal, with a difference of only 0.04 for METEOR and 0.06 for MoverScore. 
We note that the inclusion of Nucleus Sampling and greedy decoding is to provide context for the human participants, and not to directly compare against them as methods. Greedy decoding is more fluent as it produces no diversity, and Nucleus Sampling is optimized in an unrealistically favorable manner, as there is no established methodology to tune the parameter $p$ otherwise.


\begin{table}[tb]
\begin{tabular}{l|cc|cc}
        & \multicolumn{2}{c|}{Fluency}    & \multicolumn{2}{c}{Adequacy}   \\ \cline{2-5} 
        & \textit{raw} & \textit{z-score} & \textit{raw} & \textit{z-score} \\ \hline
Greedy  & \textbf{82.555}       & 0.334            & \textbf{84.233}       & 0.205            \\
IS-E   & 73.892       & 0.028            & 79.790       & 0.057            \\
IS-D  & 71.824       & -0.032           & 79.043       & 0.020            \\
IS-L    & 73.343       & -0.002           & 79.846       & 0.017            \\
Nucleus & 76.753       & 0.120            & 82.581       & 0.156           
\end{tabular}  
\caption{Human Evaluation results.}

\label{tab:human}
\end{table}

\begin{table}[tb]
\begin{tabular}{l|cc|cc}
        & \multicolumn{2}{c|}{Fluency}    & \multicolumn{2}{c}{Adequacy}   \\ \cline{2-5} 
        & \textit{raw} & \textit{z-score} & \textit{raw} & \textit{z-score} \\ \hline
IS-E   & 49.258       & -0.041           & 65.357       & 0.012            \\
IS-D  & 53.593       & 0.080            & 64.265       & -0.006           \\
IS-L    & \textbf{54.324}       & 0.104            & \textbf{65.509}       & 0.065            \\
Nucleus & 39.762       & -0.295           & 60.561       & -0.017          
\end{tabular}

\caption{Human Evaluation results for texts always sampling the last word of the reduced sample pool.}\label{tab:lastk}
\end{table}

We evaluate the fluency and adequacy of the texts via Direct Assessment \cite{graham_baldwin_moffat_zobel_2017}; a human evaluation framework that has been employed on MT \cite{bojar-etal-2018-findings}, surface realisation \cite{mille-etal-2018-first} and video captioning \cite{AwadEtAl2018} output. We used the publicly available code of Direct Assessment\footnote{https://github.com/ygraham/crowd-alone} to setup tasks on the Amazon Mechanical Turk (AMT) platform. 

To minimise correlation between the criteria, separate tasks were created asking participants to assess the fluency and adequacy of the provided texts; a 100-point Likert scale was used. For fluency, participants were asked to judge how grammatical and natural the text was. The task for adequacy was more complicated as participants were asked to compare the text with a checklist of snippets that it should include. We generated the checklist of snippets through simple rule based \textsc{nlg} (i.e. templates).
Every text was evaluated by at least 3 participants.
We limited the crowd-workers that could participate in the tasks to those residing in English-speaking countries, and who had a high acceptance rate. Even so, after consulting the participants' reliability based on the Direct Assessment platform analysis, we had to filter out 27\% and 50\% of those who assessed fluency and adequacy respectively. 

We sampled 1500 texts from each of the different Informed Sampling configurations, Nucleus Sampling and greedy inference. 
Table~\ref{tab:human} shows the raw and mean standardised z-scores of the human assessments. To determine whether the observed differences were statistically significant we used the Wilcoxon rank sum test.
On both fluency and adequacy the greedy model is the best, while IS-E and Nucleus are comparable on fluency. All other configurations have no statistically significant difference between them. This confirms that Informed sampling learns a level of diversity that the \textsc{nlg} model can generate without particularly hurting the output's quality when compared to an unrealistic optimization of Nucleus Sampling. While fluency and adequacy is lower than greedy inference (as is to be expected), the gain in diversity is significant. 

\subsection{Sample pool analysis} \label{sec:evaledge}

\begin{figure}
\centering
 \includegraphics[width=5cm]{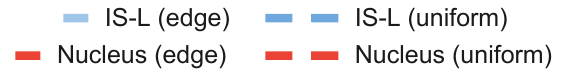} %
  \includegraphics[width=\linewidth]{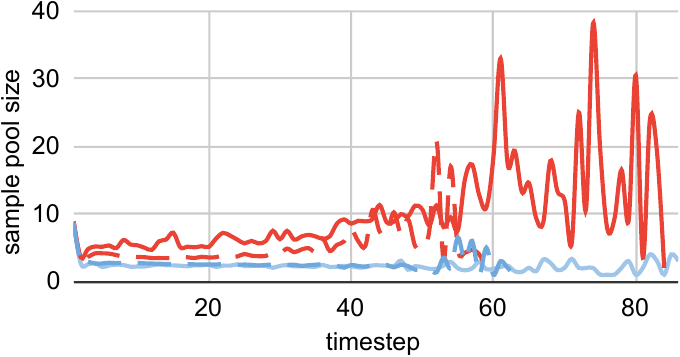}

\caption{Average sample pool size over decoding.}
  \label{fig:mean}
\end{figure}  


To better assess the edge cases of the decoding strategies, we generate 750 texts from Nucleus and each Informed Sampling configuration by always picking the least probable word in the range that each method returns. This will help us determine the quality of the texts for which the \textsc{nlg} model is least confident, but the decoding strategies still consider to be reliable enough to generate. 
Table~\ref{tab:lastk} shows the raw and mean standardised z-scores for this setting.\footnote{Automatic metrics  results are included in the appendix.}
Again, most configurations show no statistically significant difference between them, with the exception of IS-L and Nucleus on fluency. 
This shows that Informed Sampling is better at determining edge cases where it can reliably generate diverse output without hurting quality.

In addition, Figure~\ref{fig:mean} shows how the sample pool varies over the course of decoding a sentence (i.e. at each timestep) for each decoding strategy. We compare the behavior of IS-L and Nucleus, when decoding the sentences by either uniform sampling or always picking the least probable (edge) word.\footnote{IS-E and IS-D produce sample pools similar to IS-L. Full plot of Figure \ref{fig:mean} is provided in the Appendix.}
IS-L generally begins with a larger pool size at timestep $t=0$, indicating that it considers more diverse ways to begin the sentences. 
Overall, we observe that the pool size for Nucleus is larger and becomes even larger and more inconsistent at later timesteps. This is especially prevalent when picking the last word, which suggests that Nucleus leads the underlying NLG model to become less confident, possibly due to error propagation. On the other hand, IS-L demonstrates more consistent behavior, reducing its pool size over time as fewer sentence variations become available.

\section{Conclusion}

In this paper, we proposed \textit{Informed Sampling} which employs a meta-classifier exploiting diversity-specific training signals to determine which words in the decoding distribution lead to reliably diverse generation. Due to the lack of explicit training signal for diversity, we adapted three Imitation Learning frameworks and showed that their application helps \textit{Informed Sampling} determine the level of diversity that the underlying \textsc{nlg} model is comfortable to produce. Our experimental results show that \textit{Informed Sampling} leads to highly diverse output while minimising the cost to the quality of the text. We also show that \textit{Informed Sampling} is better than previous work at determining the edge cases where it can still reliably generate diverse output even though the \textsc{nlg} model assigns a lower probability.
Additionally, we presented a thorough analysis of open-domain diversity methods applied to concept-to-text \textsc{nlg}. 

\textit{Informed Sampling} is agnostic to the underlying model; its input consists of hidden states/embeddings and a probability distribution that can be obtained from almost any language generation model. In future work, we aim to extend \textit{Informed Sampling} to other language generation tasks, e.g. machine translation and open-domain \textsc{nlg}. Additionally, it would be interesting to explore the application of \textit{Informed Sampling} over the probability distribution of large pretrained models.
\bibliography{anthology,custom}
\bibliographystyle{acl_natbib}
\newpage
\appendix

\section{System Configurations} 


This section is similar to the system configurations section of the main paper, but includes many more configuration details.

Apart from the experiments with SpaceFusion \cite{gao2019jointly}, all our experiments make use of SCLSTM as underlying \textsc{nlg} model. Using the implementation provided by \citet{budzianowski2018multiwoz}\footnote{https://github.com/andy194673/nlg-sclstm-multiwoz/}, the model has been trained with 4 hidden layers, states of size 100 and 0.25 dropout, Adam \cite{kingma2015adam} as loss optimiser, learning rate of 0.005 and gradient clipping at 0.5. Early stopping was applied when validation loss did not decrease within 6 epochs. 

For our meta-classifier, we used 3 linear/ReLU layers with 512 as hidden state size, trained for 30 epochs per dataset iteration with Stochastic Gradient Descent and a learning rate of 0.05. For \textsc{lols}, exponential decay is performed with a parameter $\beta$ of 0.1. 
We initialised using a single iteration of Exact Imitation over the full dataset. 
Due to time constraints, for the following training iterations with any IL framework, only 10\% randomly selected sentences were used. We evaluated the meta-classifiers generated by our last iteration. 

In addition to our proposed method, we explored six different techniques for diversity: sampling Beam Search, First Word Control \cite{deriu2017end}, MMI-AntiLM \cite{li2016diversity}, SpaceFusion \cite{gao2019jointly}, Top-$k$ \cite{fan2018hierarchical} and Nucleus Sampling \cite{holtzman2019curious}. 
Beam Search and MMI are performed with beam size = 10 and the sentences are selected among the top 10 beams with the criteria described in Section 5.3. For the MMI objective function we implemented MMI-antiLM as suggested by \citet{li2016diversity}. For the auxiliary language model, we used a 2-layer, 650 vector size LSTM,
trained for 40 epochs on the MultiWOZ references. At decoding, the MMI-antiLM parameters were set as $\lambda=0.5$ and $g=5$.
SpaceFusion 
was trained using the configuration provided with the code. Tests on different settings did not achieve significant improvements. Values for the random vector $r$ were generated in the range $-5, 5$.
For First Word Control, we selected all the words that appear more than 60 times as first word in the training references, resulting in a set of 67 different possible first words. SCLSTM was modified as in \citet{deriu2017end} and trained with the configuration described above. At inference time, one sentence is generated per each first word and the output is selected with the criteria described in Section 5.3.
For Top-$k$ and Nucleus Sampling, since parameters $k$ and \textit{p} are not tunable, we report results for ranges 2-10 for $k$ and 0.10-0.95 for $p$. 

The aforementioned parameters in related work ($\lambda$, $g$, $r$, $k$, $p$), were all tuned based on observations of output and diversity metrics.

\section{Complete Results} \label{appendixresults}

Tables~\ref{tab:k_all_uniform} and \ref{tab:k_all_prob} show detailed results of automatic metrics for Top-\textit{k} across different values of $k$ for uniform and stochastic sampling respectively. Similarly, Tables~\ref{tab:p_all_uniform} and \ref{tab:p_all_prob} show detailed results for Nucleus Sampling across different values of $p$. These results are corresponding to those shown in Figure~3 of the main paper.

Table \ref{tab:space} shows the results for Space Fusion by varying the range of the $r$ random vector added to the input latent variable. Similarly to $k$ and $p$, performance and diversity are inversely correlated when the range is widened. However, this change does not considerably affect the slot error, which remains drastically higher than other systems.

Table~\ref{tab:lastkauto} shows automatic results for the edge case experiment presented in Section~\ref{sec:evalis} of the main paper, and correspond to the human evaluation experiments summarised in Table~\ref{tab:lastk}. Similarly to the results presented in Table \ref{tab:results}, Nucleus Sampling achieved the highest BLEU score. However, all systems performed similarly according to METEOR and MoverScore, while Informed Sampling methods produced outputs with fewer slot errors than Nucleus. Diversity metrics are not included as diversity comparison is not informative when performed on experiments where the word choice at each timestep is forced (here to the least probable word) rather than sampled.

\begin{table*}[h]
\centering
\begin{tabular}{l|ccccccc}
          & k = 1 & k = 2 & k = 3 & k = 4 & k = 5 & k = 7 & k = 10 \\ \hline
BLEU       & 0.654 & 0.336 & 0.207 & 0.144 & 0.105 & 0.069 & 0.005  \\
METEOR     & 0.496 & 0.400 & 0.368 & 0.352 & 0.343 & 0.329 & 0.253  \\
Mover      & 0.804 & 0.675 & 0.604 & 0.559 & 0.533 & 0.497 & 0.368  \\
Slot Error & 4.071 & 0.830 & 1.329 & 2.192 & 2.819 & 4.461 & 30.667 \\ \hline
1 - SB     & 0.014 & 0.093 & 0.199 & 0.297 & 0.375 & 0.483 & 0.846  \\
Dist-1     & 0.004 & 0.007 & 0.008 & 0.008 & 0.008 & 0.008 & 0.007  \\
Dist-2     & 0.022 & 0.066 & 0.096 & 0.117 & 0.134 & 0.162 & 0.400  \\
Dist-4     & 0.079 & 0.399 & 0.638 & 0.771 & 0.844 & 0.919 & 0.999  \\
Dist-Sent  & 0.307 & 0.869 & 0.943 & 0.978 & 0.985 & 0.990 & 0.996 
\end{tabular}
\caption{Complete results for \textit{Top-k} with uniform sampling.}\label{tab:k_all_uniform}
\end{table*}

\begin{table*}[h]
\centering
\begin{tabular}{l|ccccccc}
          & k = 1 & k = 2                     & k = 3                     & k = 4                     & k = 5                     & k = 7                     & k = 10                    \\ \hline
BLEU       & 0.654 & \multicolumn{1}{r}{0.633} & \multicolumn{1}{r}{0.609} & \multicolumn{1}{r}{0.594} & \multicolumn{1}{r}{0.589} & \multicolumn{1}{r}{0.578} & \multicolumn{1}{r}{0.556} \\
METEOR     & 0.496 & 0.489                     & 0.480                     & 0.476                     & 0.473                     & 0.470                     & 0.461                     \\
Mover      & 0.804 & 0.797                     & 0.784                     & 0.780                     & 0.775                     & 0.768                     & 0.752                     \\
Slot Error & 4.071 & 0.728                     & 0.652                     & 0.643                     & 0.482                     & 0.576                     & 0.559                     \\ \hline
1 - SB     & 0.014 & 0.028                     & 0.038                     & 0.047                     & 0.050                     & 0.058                     & 0.081                     \\
Dist-1     & 0.004 & 0.005                     & 0.005                     & 0.006                     & 0.006                     & 0.006                     & 0.007                     \\
Dist-2     & 0.022 & 0.032                     & 0.039                     & 0.044                     & 0.046                     & 0.051                     & 0.063                     \\
Dist-4     & 0.079 & 0.138                     & 0.177                     & 0.204                     & 0.216                     & 0.238                     & 0.289                     \\
Dist-Sent  & 0.307 & 0.592                     & 0.673                     & 0.708                     & 0.738                     & 0.770                     & 0.819                    
\end{tabular}
\caption{Complete results for \textit{Top-k} with stochastic sampling.}\label{tab:k_all_prob}
\end{table*}

\begin{table*}[h]
\centering
\resizebox{\textwidth}{!}{
\begin{tabular}{l|cccccccccccc}
          & p = 0.1 & p = 0.2 & p = 0.3 & p = 0.4 & p = 0.5 & p = 0.6 & p = 0.7 & p = 0.75 & p = 0.8 & p = 0.85 & p = 0.9 & p = 0.95 \\ \hline
BLEU       & 0.654   & 0.654   & 0.655   & 0.653   & 0.646   & 0.633   & 0.596   & 0.574    & 0.535   & 0.480    & 0.392   & 0.230    \\
METEOR     & 0.495   & 0.495   & 0.497   & 0.496   & 0.496   & 0.497   & 0.470   & 0.461    & 0.449   & 0.432    & 0.401   & 0.360    \\
Mover      & 0.804   & 0.804   & 0.804   & 0.804   & 0.799   & 0.789   & 0.766   & 0.751    & 0.7333  & 0.704    & 0.656   & 0.557    \\
Slot Error & 4.080   & 3.674   & 2.878   & 2.167   & 1.261   & 0.906   & 0.770   & 0.643    & 0.719   & 0.982     & 1.244   & 3.801    \\ \hline
1 - SB     & 0.014   & 0.014   & 0.015   & 0.018   & 0.023   & 0.030   & 0.044   & 0.057    & 0.074   & 0.109    & 0.187   & 0.398    \\
Dist-1     & 0.004   & 0.004   & 0.004   & 0.004   & 0.005   & 0.005   & 0.006   & 0.006    & 0.007   & 0.008    & 0.009   & 0.009    \\
Dist-2     & 0.022   & 0.022   & 0.023   & 0.025   & 0.030   & 0.034   & 0.042   & 0.049    & 0.058   & 0.075    & 0.113   & 0.215    \\
Dist-4     & 0.079   & 0.080   & 0.082   & 0.092   & 0.113   & 0.141   & 0.190   & 0.229    & 0.280   & 0.359    & 0.502   & 0.760    \\
Dist-Sent  & 0.307   & 0.310   & 0.331   & 0.389   & 0.480   & 0.588   & 0.727   & 0.799    & 0.863   & 0.924    & 0.957   & 0.956   
\end{tabular}}
\caption{Complete results for \textit{Nucleus Sampling} with uniform sampling.}\label{tab:p_all_uniform}
\end{table*}

\begin{table*}[h]
\centering
\resizebox{\textwidth}{!}{
\begin{tabular}{l|cccccccccccc}
          & p = 0.1 & p = 0.2 & p = 0.3 & p = 0.4 & p = 0.5 & p = 0.6 & p = 0.7 & p = 0.75 & p = 0.8 & p = 0.85 & p = 0.9 & p = 0.95 \\ \hline
BLEU       & 0.654   & 0.654   & 0.656   & 0.656   & 0.656   & 0.651   & 0.635   & 0.628    & 0.621   & 0.608    & 0.594   & 0.578    \\
METEOR     & 0.495   & 0.495   & 0.497   & 0.498   & 0.497   & 0.496   & 0.489   & 0.486    & 0.483   & 0.478    & 0.473   & 0.468    \\
Mover      & 0.804   & 0.804   & 0.805   & 0.805   & 0.806   & 0.804   & 0.794   & 0.786    & 0.790   & 0.779    & 0.772   & 0.763    \\
Slot Error & 4.088   & 3.665   & 2.920   & 2.108   & 1.405   & 0.990   & 0.813   & 0.686    & 0.686   & 0.567    & 0.626   & 0.550    \\ \hline
1 - SB     & 0.014   & 0.015   & 0.015   & 0.018   & 0.021   & 0.024   & 0.029   & 0.034    & 0.039   & 0.044    & 0.054   & 0.065    \\
Dist-1     & 0.004   & 0.004   & 0.004   & 0.004   & 0.004   & 0.005   & 0.005   & 0.006    & 0.006   & 0.006    & 0.006   & 0.007    \\
Dist-2     & 0.022   & 0.023   & 0.022   & 0.024   & 0.027   & 0.030   & 0.033   & 0.036    & 0.039   & 0.042    & 0.048   & 0.054    \\
Dist-4     & 0.079   & 0.080   & 0.081   & 0.089   & 0.100   & 0.118   & 0.139   & 0.157    & 0.170   & 0.190    & 0.216   & 0.245    \\
Dist-Sent  & 0.307   & 0.309   & 0.327   & 0.374   & 0.448   & 0.528   & 0.603   & 0.646    & 0.678   & 0.718    & 0.750   & 0.789   
\end{tabular}}
\caption{Complete results for \textit{Nucleus Sampling} with stochastic sampling.}\label{tab:p_all_prob}
\end{table*}

\begin{table*}[h]
\centering
\begin{tabular}{l|cccc}
          & $|r| = 1.5$ & $|r| = 5$  & $|r| = 10$ & $|r| = 20$ \\ \hline
BLEU       & 0.466     & 0.439  & 0.341   & 0.233   \\
METEOR     & 0.365     & 0.332  & 0.244   & 0.141   \\
Mover      & 0.671     & 0.642  & 0.537   & 0.384   \\
Slot Error & 52.98     & 45.218 & 60.344  & 83.299  \\ \hline
1 - SB     & 0.002     & 0.008  & 0.025   & 0.045   \\
Dist-1     & 0.002     & 0.003  & 0.003   & 0.004   \\
Dist-2     & 0.007     & 0.013  & 0.020   & 0.026   \\
Dist-4     & 0.019     & 0.045  & 0.099   & 0.142   \\
Dist-Sent  & 0.100     & 0.266  & 0.526   & 0.659  
\end{tabular}
\caption{Results for Space Fusion across different hypersphere radius around the latent vectors.}\label{tab:space}
\end{table*}

\begin{table*}[h]
\centering
\begin{tabular}{l|cccc}
           & \begin{tabular}[c]{@{}c@{}}Nucleus\\ $p=0.84$\end{tabular} & IS-E            & IS-D   & IS-L   \\ \hline
BLEU       & \textbf{0.243}                                           & 0.194           & 0.212  & 0.177  \\
METEOR     & 0.340                                                    & \textbf{0.350}  & 0.346  & 0.342  \\
Mover      & 0.523                                                   &   \textbf{0.567}           &  0.560      &      0.563  \\
Slot Error & 24.581                                                   & \textbf{19.325} & 19.350 & 20.112 \\ 
\end{tabular}

\caption{Automatic evaluation results for texts always sampling the last word of the reduced sample pool.}\label{tab:lastkauto}
\end{table*}


\section{Examples} \label{appendixexample}

Table \ref{tab:example_train} and \ref{tab:example_rest} show some output examples produced by each diversity method after reranking based on slot error and normalised sentence probability. We present the top 3 sentences, and do not filter out repeated sequences (as in our evaluation).

In the first example, for the meaning representation \textsc{[inform(train-ref = abc123), inform(train-price = 10)]}, all the systems generated sentences with structures similar to the greedily-decoded output. Beam Search, MMI and First Word Control behaved as described in Section \ref{sec:previous}. The rest of the systems were able to introduce some degree of diversity, while Nucleus Sampling and Space Fusion produced repeated sentences. 

On the other hand, for the \textsc{[inform(train-ref = abc123), inform(train-price = 10)]}, all the systems with the exception of Beam Search and MMI, produced diverse sentences. However, First Word Control, and Space Fusion generated some irrelevant content, while Top-$k$, Nucleus Sampling, and IS-E present some disfluency.

Table \ref{tab:example_last} illustrates some output examples generated by Greedy (as benchmark), Nucleus and Informed Sampling for the edge case experiment presented in Section~\ref{sec:evalis}. These examples correspond to the human evaluation experiments summarised in Table~\ref{tab:lastk} and Table~\ref{tab:lastkauto}. Overall, neither Informed Sampling models nor Nucleus sampling were able to generate consistently correct and fluent outputs. However, the table illustrates some examples of how catastrophic error propagation can be when non-reliable words are sampled. Specifically, for the first MR, Nucleus Sampling produced a non-sensical sentence which we attribute mainly at the generation of the tokens ``british'' and ``,''. Informed Sampling models also suffer from error propagation (as seen on the second MR), but its effects are not as frequent or severe as when using Nucleus Sampling.

\begin{table*}[h]
\centering
\resizebox{\textwidth}{!}{
\begin{tabular}{|l|l|}
\hline
\textsc{MR}                                                                              & \textsc{[inform(train-ref = abc123), inform(train-price = 10)]  }                                                                    \\ \hline
Greedy                                                                          & booking was successful , the total fee is 10 GBP payable at the station . reference number is : ABC123 .                    \\ \hline
\multirow{3}{*}{\begin{tabular}[c]{@{}l@{}}Beam\\ Search\end{tabular}}          & booking was successful , the total fee is 10 GBP payable at the station . reference number is : ABC123 .                    \\ \cline{2-2} 
                                                                                & booking was successful , the total fee is 10 GBP payable at the station . your reference number is ABC123 .                 \\ \cline{2-2} 
                                                                                & booking was successful , the total fee is 10 GBP payable at the station . reference number is : ABC123                      \\ \hline
\multirow{3}{*}{MMI}                                                            & {\color[HTML]{3531FF}booked has 10 GBP thing} . reference number is : ABC123 .                                                                    \\ \cline{2-2} 
                                                                                & {\color[HTML]{3531FF}booked has mention fee} of 10 GBP . reference number is : ABC123 .                                                           \\ \cline{2-2} 
                                                                                & {\color[HTML]{3531FF}booked has mention total fee is} 10 GBP payable at the station . reference number is : ABC123 .                              \\ \hline
\multirow{3}{*}{\begin{tabular}[c]{@{}l@{}}First\\ Word\\ Control\end{tabular}} & booking was successful , the total fee is 10 GBP payable at the station . reference number is : ABC123 .                    \\ \cline{2-2} 
                                                                                & ok. the booking was successful , the total fee is 10 GBP payable at the station . reference number is : ABC123 .            \\ \cline{2-2} 
                                                                                & your booking was successful , the total fee is 10 GBP payable at the station . reference number is : ABC123 .               \\ \hline
\multirow{3}{*}{\begin{tabular}[c]{@{}l@{}}Space\\ Fusion \\$|r| = 5$\end{tabular}}         & booking was successful , the total fee is 10 GBP payable at the station . reference number is : ABC123 .                    \\ \cline{2-2} 
                                                                                & booking was successful , the total fee is 10 GBP payable at the station . reference number is : ABC123 .                     \\ \cline{2-2} 
                                                                                & the booking was successful , the total fee is 10 GBP payable at the station . reference number is : ABC123 .                 \\ \hline
\multirow{3}{*}{\begin{tabular}[c]{@{}l@{}}Top-\textit{k} \\$k = 2$\end{tabular}}                                                          & booking is complete . your reference number is ABC123 and it will be 10 GBP .                                               \\ \cline{2-2} 
                                                                                & {\color[HTML]{3531FF}your train has booked} . your total fee is 10 GBP and your reference number is ABC123                                        \\ \cline{2-2} 
                                                                                & booking was completed . the reference is ABC123 and it will cost 10 GBP .                                                   \\ \hline
\multirow{3}{*}{\begin{tabular}[c]{@{}l@{}}Nucleus\\ Sampling\\$p = 0.84$\end{tabular}}     & booking was successful , the total fee is 10 GBP payable at the station . reference number is : ABC123 .                    \\ \cline{2-2} 
                                                                                & booking was successful , the total fee is 10 GBP payable at the station . reference number is : ABC123 .                    \\ \cline{2-2} 
                                                                                & the total is 10 GBP and your reference number is ABC123 .                    \\ \hline
\multirow{3}{*}{IS-E}    & yes , the booking was successful . your reference number is : ABC123 , the cost is 10 GBP .                                 \\ \cline{2-2} 
                                                                                & yes , the booking was successful , you reference number is ABC123 . 10 GBP payable at the station                           \\ \cline{2-2} 
                                                                                & your tickets have been reserved . your total is 10 GBP , which you can pay at the station , your reference \# is ABC123 \\ \hline
\multirow{3}{*}{IS-D}                                                         & booking was successful . reference number is 10 GBP payable at the station . your reference number is : ABC123 .            \\ \cline{2-2} 
                                                                                & your ticket has been booked ! your reference number is : ABC123 , the price is 10 GBP .                                     \\ \cline{2-2} 
                                                                                & yes , your tickets have been booked ! the  {\color[HTML]{FE0000}cost} is ABC123 and the total cost is 10 GBP .                                \\ \hline
\multirow{3}{*}{IS-L}                                                           & great ! booking was successful and the fee is 10 GBP which you can pay at the station . your reference number is : ABC123 . \\ \cline{2-2} 
                                                                                & yes , booking was successful and your total is 10 GBP . you can pay that at the station . reference number is ABC123        \\ \cline{2-2} 
                                                                                & great ! booking was successful ! your reference is ABC123 and you will pay 10 GBP at the station .                          \\ \hline
\end{tabular}
}
\caption{Top 3 outputs for the \textsc{MR} \textsc{[inform(train-ref = abc123), inform(train-price = 10)]}. {\color[HTML]{FE0000}\textsc{Red}} and {\color[HTML]{3531FF}\textsc{blue}} text mark adequacy and fluency errors respectively.}\label{tab:example_train}
\end{table*}

\begin{table*}[h]
\centering
\resizebox{\textwidth}{!}{
\begin{tabular}{|l|l|}
\hline
\textsc{MR}                                                                              &   
\textsc{[inform(rest-name = Tandori), inform(rest-area = East)]    } 

\\ \hline
Greedy                                                                          & Tandoori is in the East side . would you like me to book it for you ?                                                                                                                                          \\ \hline
\multirow{3}{*}{\begin{tabular}[c]{@{}l@{}}Beam\\ Search\end{tabular}}          & Tandoori is located in the East side . would you like me to book it for you ?                                                                                                                                  \\ \cline{2-2} 
                                                                                & Tandoori is in the East side . would you like me to book it for you ?                                                                                                                                          \\ \cline{2-2} 
                                                                                & Tandoori is located in the East side . would you like to book a table ?                                                                                                                                        \\ \hline
\multirow{3}{*}{MMI}                                                            & Tandoori  {\color[HTML]{3531FF}going} is located in the East side                                                                                                                                                                     \\ \cline{2-2} 
                                                                                & Tandoori {\color[HTML]{3531FF}going} is located in the East side .                                                                                                                                                                   \\ \cline{2-2} 
                                                                                & Tandoori {\color[HTML]{3531FF}going} is located in the East side . would you like a reservation ?                                                                                                                                    \\ \hline
\multirow{3}{*}{\begin{tabular}[c]{@{}l@{}}First\\ Word\\ Control\end{tabular}} & unfortunately , i have {\color[HTML]{FE0000}4 Indian restaurants} . i have Tandoori {\color[HTML]{FE0000} in the South and 1 in the  West , and 1} in the  East .                                                                                           \\ \cline{2-2} 
                                                                                & yes , there are {\color[HTML]{FE0000}4 Indian restaurants}  . Tandoori in  {\color[HTML]{FE0000} in the South and 1 in the  West , and 1} in the  East .                                                                                                         \\ \cline{2-2} 
                                                                                & of course ! there are {\color[HTML]{FE0000}4 Indian restaurants in the South}  , Tandoori {\color[HTML]{FE0000}in the  West and 1} in the  East .                                                                                                           \\ \hline
\multirow{3}{*}{\begin{tabular}[c]{@{}l@{}}Space\\ Fusion\\$|r| = 5$\end{tabular}}         & Tandori is in East .                                                                                                                                                                                           \\ \cline{2-2} 
                                                                                & \begin{tabular}[c]{@{}l@{}}i can book you a table for you . {\color[HTML]{FE0000}i can book you a table for you . how many people and for how many nights and} \\ {\color[HTML]{FE0000}how many nights and how many people will be staying ?}\end{tabular} \\ \cline{2-2} 
                                                                                & i have the Tandori in the East .                                                                                                                                                                               \\ \hline
\multirow{3}{*}{\begin{tabular}[c]{@{}l@{}}Top-$k$\\$k=2$\end{tabular}}                                                          & i have the Tandoori in East side , shall i book you ?                                                                                                                                                          \\ \cline{2-2} 
                                                                                & i have a listing in Tandoori , it is in East side , can i make a booking ?                                                                                                                                     \\ \cline{2-2} 
                                                                                & i have a restaurant {\color[HTML]{3531FF}located at} Tandoori in East side , can i make you a reservation {\color[HTML]{3531FF}at the address of them}?                                                                                                   \\ \hline
\multirow{3}{*}{\begin{tabular}[c]{@{}l@{}}Nucleus\\ Sampling\\$p = 0.84$\end{tabular}}     & Tandoori is in the East side part of town . would you like a reservation ?                                                                                                                                     \\ \cline{2-2} 
                                                                                & it 's located in the East side . i can {\color[HTML]{3531FF}book a reservation} at Tandoori if you are interested .                                                                                                                  \\ \cline{2-2} 
                                                                                & would you like to book a table at Tandoori at the East side ,                                                                                                                                                  \\ \hline
\multirow{3}{*}{IS-E}      & there is a restaurant in the East side that meets those criteria . would you like to try Tandoori ?                                                                                                            \\ \cline{2-2} 
                                                                                & Tandoori is in the East side {\color[HTML]{3531FF}part} of the city . would you like me to check availability ?                                                                                                                      \\ \cline{2-2} 
                                                                                & i have a listing in the East side . would you like {\color[HTML]{3531FF}to book you a reservation} at the Tandoori ?                                                                                                                 \\ \hline
\multirow{3}{*}{IS-D}                                                         & Tandoori is a nice restaurant in the East side                                                                                                                                                                 \\ \cline{2-2} 
                                                                                & i can book Tandoori in the East side , would you like reservations ?                                                                                                                                           \\ \cline{2-2} 
                                                                                & Tandoori is in the East side area , shall i book your table ?                                                                                                                                                  \\ \hline
\multirow{3}{*}{IS-L}                                                           & i have a listing for the restaurant in the East side . would you like me to make a reservation at Tandoori for you ?                                                                                           \\ \cline{2-2} 
                                                                                & Tandoori is in the East side side of the city . would you like the reservation ?                                                                                                                               \\ \cline{2-2} 
                                                                                & Tandoori is an excellent restaurant in the East side . shall i make the reservation ?                                                                                                                          \\ \hline
\end{tabular}
}
\caption{Top 3 outputs for the \textsc{MR} \textsc{[inform(rest-name = Tandori), inform(rest-area = East)]}.   {\color[HTML]{FE0000}\textsc{Red}} and {\color[HTML]{3531FF}\textsc{blue}} text mark adequacy and fluency errors respectively.}\label{tab:example_rest}
\end{table*}

\begin{table*}[h]
\centering
\resizebox{\textwidth}{!}{\begin{tabular}{|l|l|}
\hline
\textsc{MR}      &\textsc{[inform(restaurant-price = Expensive)]}                                                                                                                                                                            \\ \hline
Greedy  & it is in the expensive price range .                                                                                                                                                                            \\ \hline
Nucleus & \begin{tabular}[c]{@{}l@{}}great the british chinese expensive options , does n't matter when you have it there and just for additional \\ info and welcome again if you 're interested in books .\end{tabular} \\ \hline
\textsc{IS-E}    & sure . they 're in a expensive price .                                                                                                                                                                          \\ \hline
\textsc{IS-D}    & its in a expensive price .                                                                                                                                                                                      \\ \hline
\textsc{IS-L}      & the restaurant is expensive .                                                                                                                                                                                   \\ \hline
\textsc{MR}        &\textsc{[inform(attraction-name = Jello Gallery; Kettles Yard)]}                                                                                                                            \\ \hline
Greedy  & cafe jello gallery and  kettles yard are both great options .                                                                                                                                                   \\ \hline
Nucleus & well and the cafe jello gallery on my favorites i were listed are looking for you either of the .                                                                                                               \\ \hline
\textsc{IS-E}    & sure ! i can suggest the cafe jello gallery ,  kettles yard .                                                                                                                                                   \\ \hline
\textsc{IS-D}    & sure , i would like to suggest cafe jello gallery .                                                                                                                                                             \\ \hline
\textsc{IS-L}    & there 's cafe jello gallery or the  kettles yard , or slot-attraction-recommend-name .                                                                                                                          \\ \hline
\textsc{MR}        & \textsc{[request(booking-day; booking-people; booking-stay]}                                                                                                          \\ \hline
Greedy  & \begin{tabular}[c]{@{}l@{}}i can help you with that . how many people will be staying , and what day will you be arriving , and \\ how many nights will you be staying ?\end{tabular}                           \\ \hline
Nucleus & and just what day and time how many people is it staying for you ?                                                                                                                                              \\ \hline
\textsc{IS-E}    & okay , can you give me the details ?                                                                                                                                                                            \\ \hline
\textsc{IS-D}      & okay , can you give me the details ?                                                                                                                                                                            \\ \hline
\textsc{IS-L}     & \begin{tabular}[c]{@{}l@{}}how long is your stay ? what day do you plan on arriving , for how long are you staying and what time\\ would like your reservation to book the restaurant for you ?\end{tabular}    \\ \hline
\end{tabular}
}
\caption{Output text generated by Nucleus, Informed Sampling and Greedy on sample pool edge cases.}\label{tab:example_last}
\end{table*}

\begin{figure*}
\centering
 \includegraphics[width=7cm]{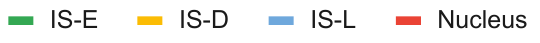} %
  \includegraphics[width=.85\linewidth]{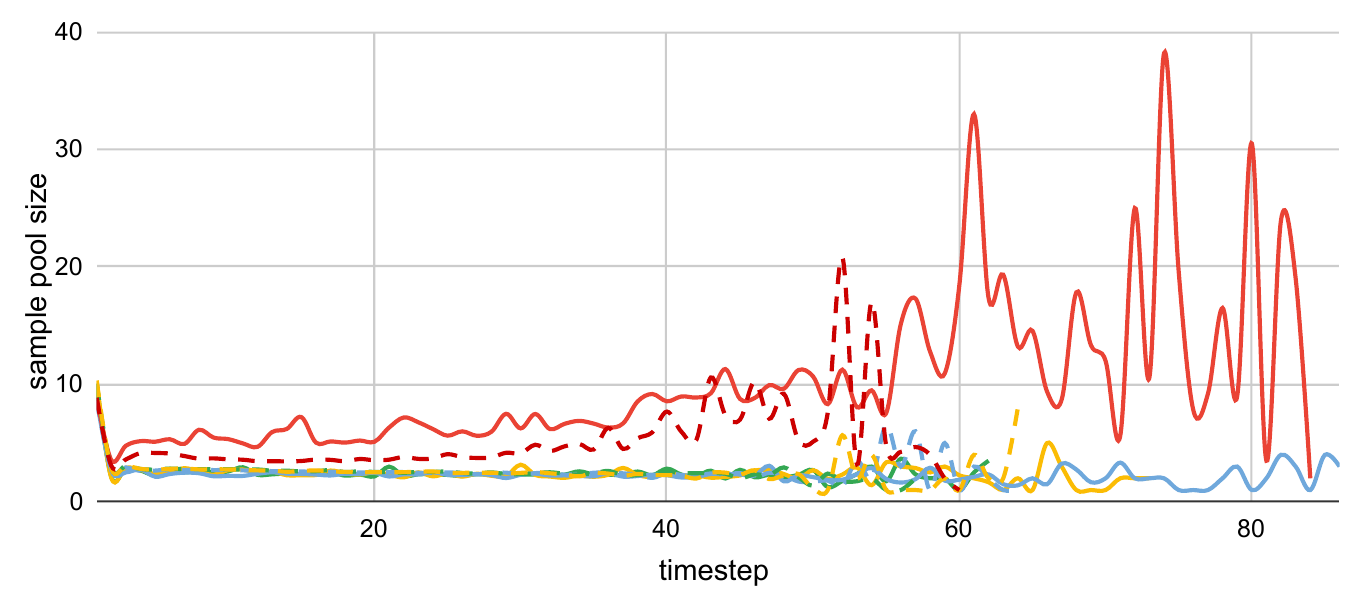}

\caption{Average sample pool size over decoding. Solid lines correspond to uniform sampling, and dashed lines correspond to sampling the least probable word in the sample (i.e. the edge case experiment).}
  \label{fig:mean2}
\end{figure*}  

\section{Human evaluation platform examples} \label{appendix_eval}

\begin{figure*}[h!]
\centering
  \includegraphics[width=\linewidth]{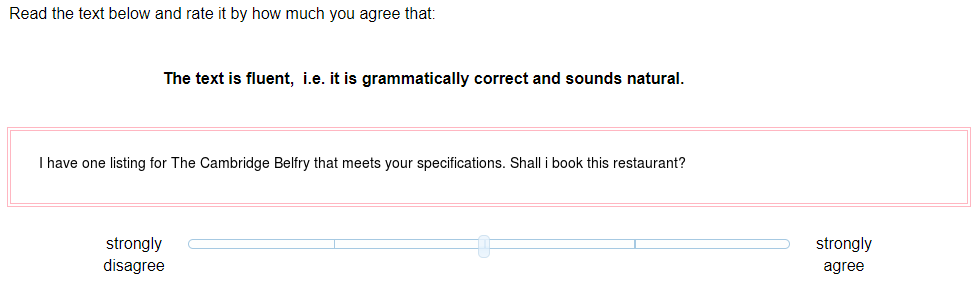}
\caption{Evaluation platform for assessment of output fluency.}
  \label{fig:eval_platform_fluency}
\end{figure*}

\begin{figure*}[t!]
\centering
  \includegraphics[width=\linewidth]{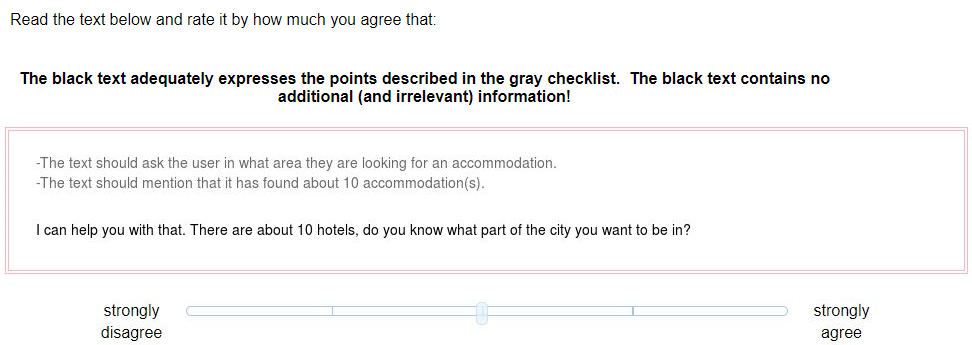}
\caption{Evaluation platform for assessment of output adequacy.}
  \label{fig:eval_platform_adequacy}
\vspace{128in}
\end{figure*}

Figures \ref{fig:eval_platform_fluency} and \ref{fig:eval_platform_adequacy} show examples of the evaluation platform as shown to the human participants of Amazon Mechanical Turn. Figure~\ref{fig:eval_platform_fluency} asks the participants to rate the fluency of the text, while Figure~\ref{fig:eval_platform_adequacy} is used to rate adequacy. For the latter, people were asked to compare the text with a checklist of snippets that it should include. The checklist of snippets was generated through simple rule based \textsc{nlg} (i.e. manually authored templates).

\end{document}